\documentclass[11pt]{article}
\usepackage{pdfpages}
\usepackage{times}
\usepackage{hyperref}
\usepackage{amsmath, amssymb, amstext, amsbsy}
\usepackage{booktabs}
\usepackage{fullpage}
\usepackage{mathtools}
\usepackage{natbib}
\usepackage{caption}
\usepackage{subcaption}
\usepackage{graphicx}

\newcommand{\bone}{\boldsymbol{1}}

\def\Var{\mathop{\rm Var}\nolimits}

\def\diag{\mathop{\rm diag}\nolimits}

\newcommand{\bu}{\boldsymbol{u}}
\newcommand{\bv}{\boldsymbol{v}}

\newcommand{\by}{\boldsymbol{y}}

\newcommand{\bX}{\boldsymbol{X}}

\newcommand{\bbeta}{\boldsymbol{\beta}}

\newcommand{\bepsilon}{\boldsymbol{\epsilon}}

\DeclareMathOperator{\supp}{supp}

\begin{document}

\title{Iterative Hard Thresholding for Model Selection in Genome-Wide Association Studies}
\author{
    Kevin L. Keys\,$^{\text{\textbf{1}}*}$,
    Gary K. Chen\,$^{\text{\textbf{2}}}$,
    and Kenneth Lange\,$^{\text{\textbf{3}}}$
    \\
    $^{\text{\sf 1}}$ Department of Medicine, University of California, San Francisco, Box 2911, San Francisco, CA 94158 \\ 
    $^{\text{\sf 2}}$ Division of Biostatistics, University of Southern California, Los Angeles, CA 90089 \\
    $^{\text{\sf 3}}$ Departments of Biomathematics, Human Genetics, and Statistics, \\ University of California, Los Angeles, CA 90095 \\
    $^\ast$To whom correspondence should be addressed. \\
    \textbf{Email}: \texttt{kevin.keys@ucsf.edu} \\
    \textbf{Phone}: +1 (415) 514-9929
}
%\usepackage{fancyhdr}
%\pagestyle{fancy}
%\lhead{K.L. Keys, G.K. Chen, K. Lange}
%\lhead{}
%\rhead{IHT for GWAS}

\maketitle 

\abstract{
A genome-wide association study (GWAS) correlates marker and trait variation
in a study sample. Each subject is genotyped at a multitude
of SNPs (single nucleotide polymorphisms) spanning the genome. 
Here we assume that subjects are randomly collected unrelateds and that trait values are normally distributed or can be transformed to normality. 
Over the past decade, geneticists have been remarkably successful in applying GWAS analysis to hundreds of traits. 
The massive amount of data produced in these studies presents unique computational challenges. 
Penalized regression with LASSO or MCP penalties is capable of selecting a handful of associated SNPs from millions of potential SNPs. 
Unfortunately, model selection can be corrupted by false positives and false negatives, obscuring the genetic underpinning of a trait.
Here we compare LASSO and MCP penalized regression to iterative hard thresholding (IHT).
On GWAS regression data, IHT is better at model selection
and comparable in speed to both methods of penalized regression.
This conclusion holds for both simulated and real GWAS data.
IHT fosters parallelization and scales well in problems with large numbers of causal markers.
Our parallel implementation of IHT accommodates SNP genotype compression and exploits multiple CPU cores and graphics processing units (GPUs). 
This allows statistical geneticists to leverage commodity desktop computers in GWAS analysis and to avoid supercomputing.\\
\textbf{Availability:} Source code is freely available at \texttt{https://github.com/klkeys/IHT.jl}.
}

\section{Introduction}

Over the past decade, genome-wide association studies (GWASs) have benefitted from technological advances in dense genotyping arrays, high-throughput sequencing, and more powerful computing resources. 
Yet researchers still struggle to find the genetic variants that account for the missing heritability of many traits. 
It is now common for consortia studying a complex trait such as height to pool results across multiple sites and countries. 
Meta-analyses have discovered hundreds of statistically significant SNPs, each of which explains a small fraction of the total heritability. 
A drawback of GWAS meta-analysis is that it relies on univariate regression rather than on more informative multivariate regression \cite{yangetal2010}. 
Because the number of SNPs (predictors) in a GWAS vastly exceeds the number of study subjects (observations), statistical geneticists have resorted to machine learning techniques such as penalized regression \cite{langepappsinsheimersobel2014} for model selection.

In the statistical setting of $n$ subjects and $p$ predictors with $n \ll p$,
penalized regression estimates a sparse parameter vector $\bbeta \in \mathsf{R}^{p}$ by minimizing an appropriate objective function $f(\bbeta)+\lambda p(\bbeta)$,
where $f(\bbeta)$ is a convex loss, $p(\bbeta)$ is a suitable penalty, 
and $\lambda$ is a tuning constant controlling the sparsity of $\bbeta$.
The most popular and mature sparse regression tool is LASSO ($\ell_1$) regression \cite{chen1994basis,tibshirani1996}.
Unfortunately, LASSO parameter estimates are biased towards zero \cite{ElemStatLearn},
usually severely so. As a consequence of shrinkage,  LASSO regression lets too many false positives enter a model. Since GWAS is often followed by expensive biological validation studies, there is value in reducing false positive rates. 
To counteract the side effects of shrinkage, Zhang \cite{zhang2010} recommends the minimax concave penalty (MCP) as an alternative to the $\ell_1$ penalty. 
Other non-convex penalties exist, but MCP is probably the simplest to implement.
MCP also has provable convergence guarantees. 
In contrast to the LASSO, which admits too many false positives, MCP tends to allow too few predictors to enter a model. 
Thus, its false negative rate is too high. 
Our subsequent numerical examples confirm these tendencies.

Surprisingly few software packages implement efficient penalized regression algorithms for GWAS. 
The R packages \texttt{glmnet} and \texttt{ncvreg} are ideal candidates, given their ease of use, maturity of development, and wide acceptance. 
The former implements LASSO-penalized regression \cite{friedmanetal2010, lange2010numerical,tibshirani1996}, while the latter implements both LASSO- and MCP-penalized regression \cite{brehenyhuang2011, zhang2010}. 
Both packages provide excellent functionality for moderately sized problems. 
However, R's poor memory management hinders the scalability of both algorithms. 
In fact, analysis on a typical workstation is limited to at most a handful of chromosomes at a time. 
Larger problems must appeal to cluster or cloud computing. 
Neither \texttt{glmnet} nor \texttt{ncvreg} natively support the compressed PLINK binary genotype file (BED file) format so effective in storing and distributing  GWAS data \cite{purcelletal2007}. 
Scalable implementations of LASSO for GWAS with PLINK files appear in the packages Mendel, \texttt{gpu-lasso}, SparSNP, and the beta version of PLINK 1.9 \cite{abrahametal2012, changetal2015, chen2012, langeetal2013, wulange2008}. 
All of these packages include parallel computing capabilities for large GWAS datasets. Mendel, \texttt{gpu-lasso}, and PLINK 1.9 beta have multicore acceleration capabilities, while SparSNP can be run on compute clusters.
To our knowledge, only Mendel supports MCP regression with PLINK files.

As an alternative to penalized regression, one can tackle sparsity directly through greedy algorithms for sparse reconstruction.
The matching pursuit (MP) \cite{mallat1993matching} algorithm from the signals processing literature reconstructs a signal by adding predictors piecemeal, eventually yielding a sparse representation of the signal. 
This is a generalization of the older statistical procedure of forward stagewise regression \cite{donoho2012sparse}.
Similar algorithms treated in the signal processing literature include hard thresholding pursuit (HTP) \cite{bahmanirajboufounos2013, foucart2011, yuanlizhang2013}, 
orthogonal matching pursuit (OMP) \cite{pati1993orthogonal, troppgilbert2007},
compresive sample matching pursuit (CoSaMP) \cite{needelltropp2009}
and subspace pursuit (SP) \cite{daimilenkovic2009}. 

An alternative approach is to handle sparsity through projection onto sparsity sets \cite{blumensath2012, blumensathdavies2008, blumensathdavies2009, blumensathdavies2010}. 
Iterative hard thresholding (IHT) minimizes a loss function $f(\bbeta)$ subject to the sparsity constraint $\|\bbeta\|_0 \le k$,
where the $\ell_0$ ``norm" $\|\bbeta\|_0$ counts the number of nonzero entries of the parameter vector $\bbeta$.
The integer $k$ serves as a tuning constant analogous to $\lambda$ in LASSO and MCP regression. 
IHT can be viewed as a version of projected gradient descent tailored to sparse regression. 
Since these algorithms rely solely on gradients, they avoid computing and inverting large Hessian matrices and hence scale well to large datasets.

Like matching pursuit algorithms, IHT is fundamentally a greedy selection procedure.
Distinguishing which greedy algorithm demonstrates superior performance is no simple task.
Typical performance metrics include computational speed, signal recovery behavior, and convergence guarantees in noisy environments.
Although from a theoretical point of view, no greedy algorithm is uniformly superior to the others, IHT demonstrates the best balance of these three criteria among greedy algorithms \cite{blanchard2011phase, blanchard2015performance}.
Implementation details such as memory management, hardware platform, and choice of computing environment can complicate this picture.
In light of established results with greedy algorithms, we believe that a careful implementation of IHT can provide sparse approximation performance that is competitive or superior to current penalized regression procedures.

Our implementation of IHT addresses some of the specific concerns of GWAS. 
First, it accommodates genotypes presented in compressed PLINK format.
Second, our version of IHT allows the user to choose the sparsity level $k$ of a model.  
In contrast, LASSO and MCP penalized regression must choose model size indirectly by adjusting the tuning constant $\lambda$ to match a given $k$.
Third, our version of IHT is implemented in the package \texttt{IHT.jl} in the free Julia programming language.
Julia works on a variety of hardware platforms, 
encourages prudent control of memory, exploits all available CPU cores, and interfaces with massively parallel graphics processing unit (GPU) devices. 
Finally, IHT performs more precise model selection than either LASSO or MCP penalized regression. 
    We use ``precise'' in the information theoretic sense: given the sets of markers selected by each of the three algorithm, the set that IHT selects contains a higher proportion of causal markers than those of LASSO and MCP.
While this does not mean that IHT consistently recovers all causal markers, the markers it does recover are more credible than the markers that
LASSO and MCP recover.
All of these advantages can be realized on a modern desktop computer. 
Although our current IHT implementation is limited to ordinary linear least squares, the literature suggests that logistic regression is within reach \cite{bahmanirajboufounos2013, yuanlizhang2013}. 

It worth stressing that our focus is on parameter estimation and model selection.
Historically IHT lacked a coherent inference framework for constructing valid post-selection confidence intervals and $P$-values.
A recent paper \cite{yang2016selective} tries to fill this gap for group IHT; its applicability to this work is tenuous.
Post-selection inference theory for the LASSO \cite{lockhartetal2014,taylortibshirani2015,lee2016exact} is implemented in the R package \texttt{selectiveInference}. 
Because this package lacks PLINK file support and parallel processing capabilities, its scalability to GWAS is problematic. 

Before moving onto the rest of the paper, let us sketch its main contents. Section \ref{sec:methods} describes penalized regression and greedy algorithms, including IHT.
We dwell in particular on the tactics necessary for parallelization of IHT. Section \ref{sec:results} records our numerical experiments.
The performance of IHT, LASSO, and MCP regression algorithms is evaluated by several metrics: computation time, precision, recall, prediction error, and heritability.
The sparsity level $k$ for a given dataset is chosen by cross-validation on both real and simulated genetic data.
Our discussion in Section \ref{sec:discussion} summarizes results, limitations, and precautions.

\section{Methods}
\label{sec:methods}

\subsection{Penalized regression}
\label{sec:penalized_regression}

Consider a statistical design matrix $\bX \in \mathsf{R}^{n \times p}$,
a noisy $n$-dimensional response vector $\by$, and a sparse parameter vector $\bbeta$
of regression coefficients. When $\by$ represents a continuous phenotype,
then the residual sum of squares loss
\begin{equation} \label{eq:leastsquaresloss}
	f(\bbeta ) = \frac{1}{2} \left\| \by - \bX \bbeta \right\|_2^2
\end{equation} 
is appropriate for a linear model $\by = \bX \bbeta^{*} + \bepsilon$ with a Gaussian error
vector $\bepsilon$ with independent components. The goal of penalized regression is to recover the true vector $\bbeta_{*}$ of regression coefficients.

LASSO penalized regression imposes the convex $\ell_1$ penalty
$p_{\lambda}(\bbeta) = \lambda \| \bbeta \|_1 = \lambda \sum_{i = 2}^{p} | \beta_i |$. In most applications,
the intercept contribution $|\beta_1|$ is omitted from the penalty.
Various approaches exist to minimize the objective $f(\bbeta)+\lambda \|\bbeta\|_1$,
including least angle regression (LARS) \cite{efronetal2004}, 
cyclic coordinate descent \cite{friedmanetal2007, wu2009genome, wulange2008},
and the fast iterative shrinkage and thresholding algorithm (FISTA) \cite{beckteboulle2009}.
The $\ell_1$ norm penalty induces both sparsity and shrinkage.
Shrinkage per se is not an issue because
selected parameters can be re-estimated omitting the non-selected parameters and the penalty.
However, the severe shrinkage induced by the LASSO inflates false positive 
rates. In effect, spurious predictors enter the model to absorb the unexplained variance
left by the shrinkage imposed on the true predictors.

The MCP penalty takes the form $p_{\lambda, \gamma}(\bbeta) = \sum_{i=2}^p q(|\beta_i|)$
with 
\begin{align}
	\label{eq:mcp}
	q(\beta_i) & = \begin{cases} \lambda \beta_i - \beta_i^2 / (2 \gamma) 
	& 0 \le \beta_i \leq \gamma \lambda \\ \gamma \lambda^2 / 2 & \beta_i > \gamma \lambda \end{cases} \nonumber \\
	q'(\beta_i) & = \begin{cases} \lambda - \beta_i / \gamma & 0 \le \beta_i < \gamma \lambda \\ 0 & \beta_i > \gamma \lambda \end{cases}
\end{align}
for positive tuning constants $\lambda$ and $\gamma$.
The penalty \eqref{eq:mcp} attenuates penalization for large parameter values.
Indeed, beyond $\beta_i = \gamma \lambda$, MCP does not subject $\beta_i$ to further shrinkage. Relaxing penalization of large entries of $\bbeta$ ameliorates LASSO's shrinkage. If one majorizes the MCP function $q(\beta_i)$ by a scaled
absolute value function, then cyclic coordinate descent parameter updates resemble the corresponding LASSO updates \cite{jianghuang2011}.

\subsection{Greedy pursuit algorithms}

The $\ell_1$ penalty is the smallest convex relaxation of the $\ell_0$ penalty.
As mentioned earlier, one can obtain sparsity without shrinkage by directly minimizing $f(\bbeta)$ subject to $\|\bbeta\|_0 \le k$.
This subset selection problem is known to be NP-hard \cite{golubklemastewart1976, natarajan1995}.
Greedy pursuit algorithms (MP, OMP, CoSaMP, SP, HTP, IHT) can at best approximate the 
solution of the subset selection problem.
Here we sketch the main idea of each approach and describe some subtle differences between them.

MP and OMP, which build $\bbeta$ stagewise, are easy to describe.
At stage $k$ with reduced predictor matrix $\bX_{k}$ and reduced parameter vector $\bbeta_k$,
OMP computes the least squares solution 
\begin{align}
    \bbeta_k & = \bX^{+}_k\by 
     = (\bX_k^T \bX_k)^{-1} \bX_k^T \by
   \label{eq:normal_equations}
\end{align}
to the normal equations, where $\bX^{+}_k$ denotes the pseudoinverse of  $\bX_k$. Note that
$\bbeta_{k}$ can be computed algebraically when $k$ is small or iteratively when $k$ is large.
The next predictor to add is determined by the largest entry in magnitude of the gradient
\begin{align}
\nabla f(\bbeta) & = -\bX^T (\by - \bX \bbeta) 
= -\bX^T (\by - \bX_k \bbeta_k). \label{normal_grad}
\end{align}
The main difference between MP and OMP is that OMP re-estimates all currently selected
regression coefficients once a new predictor is added. In contrast, MP fixes a regression coefficient
once it is estimated. The more complex strategy of OMP gives it a slight edge in recovery performance at the cost of additional computation.

One potentially detrimental feature of OMP is that indices added to the active set remain there.
CoSaMP and SP extend OMP by adding a debiasing step, thus permitting predictors to enter and exit during the model building process. At stage $k$ debiasing is accomplished by taking the
estimate $\bbeta_k$ derived from equation (\ref{eq:normal_equations}), computing its gradient
$\nabla f(\bbeta_k)$, and identifying the $k$ largest components in magnitude of 
$\nabla f(\bbeta_k)$. The identified components are then appended to the nonzero components of 
$\bbeta_k$, and all $2k$ components are refit.  
The largest $k$ components of $\bbeta_{2k}$ in magnitude then populate the revised $k$-sparse approximation $\bbeta_{k}$. Once debiasing is complete, the sparsity level $k$ is increased to 
$k+1$, and the process is repeated. 

IHT and HTP approximate the solution to the subset selection problem
by taking the projected gradient steps
\begin{equation}
\label{eq:ihta}
	\bbeta^{+} = P_{S_{k}} [ \bbeta - \mu \nabla ( \bbeta ) ],
\end{equation}
where $\mu$ denotes the step size of the algorithm, and $P_{S_{k}}(\bbeta)$ denotes the projection of $\bbeta$ onto the sparsity set $S_{k}$ where at most $k$ components of a vector are nonzero. 
For sufficiently small $\mu$, the projected gradient update \eqref{eq:ihta} is guaranteed to reduce the loss, but it forfeits stronger convergence properties because $S_{k}$ is nonconvex.
Projection is achieved by setting all but the $k$ largest components of $\bbeta$ in magnitude equal to 0.
HTP projects by solving the normal equations in the form \eqref{eq:normal_equations} on the active set $\supp(\bbeta_{k})$. 

The pure gradient nature of IHT explains its the speed and scalability advantages over other greedy algorithms as $k$ grows.
Although the method of conjugate gradients can quickly compute the solution vector \eqref{eq:normal_equations} when $k$ is large and $\bX$ is sparse,
typical GWAS datasets involve dense predictor matrices $\bX$.
Direct solution of the normal equations then has computational complexity $\mathcal{O}(k^{3})$.
In contrast, the projection $P_{S_{k}}$ in IHT succumbs to fast sorting algorithms with computational complexity of just $\mathcal{O}(k \log p)$.
For small $k$, IHT is not intrinsically faster than other greedy pursuit algorithms, but the performance gap increases quickly as $k$ grows.
This advantage is particularly relevant in heritability estimation since many complex traits depend on hundreds or thousands of SNPs with small individual effect. 

\subsection{Convergence of IHT}

Convergence guarantees for IHT revolve around three criteria. 
Let $\bbeta_{*}$ denote the parameter vector under the true model, and let $\bbeta_{k}$ be the current estimate of $\bbeta_{*}$. Convergence guarantees consider any or all of the following quantities: (a) $\| f(\bbeta_{k}) - f(\bbeta_{*}) \|_2$, (b) $\| \bbeta_{k} - \bbeta_{*} \|_2$,  
and (c) $\| \supp(\bbeta_{k}) - \supp(\bbeta_{*})\|_1$, where $\supp(\bbeta)$ denotes a 0/1
vector conveying the support of $\bbeta$. Convergence criteria (a) and (b) are better understood than criterion (c), so we first focus on (a) and (b).

The original convergence guarantees for IHT \cite{blumensathdavies2008, blumensathdavies2010} relied on the \emph{restricted isometry property} (RIP) \cite{candesrombergtao2006} and the \emph{mutual coherence property} \cite{donoho2001uncertainty,tropp2006just} to show that criteria (a) and (b)
converge to 0.  RIP and mutual coherence together require that the normalized version of $\bX$ approximate an orthonormal matrix whose columns are uncorrelated.
However, RIP offers pessimistic worst-case bounds. 
Recent research has derived realistic guarantees of stable convergence and model recovery for HTP and by extension to to IHT. 
A combination of \emph{restricted strong convexity} (RSC) \cite{dobsonbarnett2008, loh2015regularized} and \emph{restricted strong smoothness} (RSS) \cite{agarwalnegahbanwainwright2012, jain2014iterative} places local bounds on the curvature of the loss function.
If $f(\bbeta)$ satisfies RSC and RSS, then HTP converges to a $k$-sparse minimizer $\bbeta$  provided the extreme eigenvalues of the Hessian matrix $\nabla^2 f(\bbeta_\sharp)$ are bounded for any $k$-sparse approximation $\bbeta_\sharp$ near $\bbeta$. 
The adaptation of RSC and RSS to IHT was made by Bahmani, Raj, and Boufounos \cite{bahmanirajboufounos2013}.
They invoke the \emph{stable restricted Hessian} (SRH) and the \emph{stable restricted linearization} (SRL) conditions to bound the curvature of $f(\bbeta)$ over a restricted subset of the domain. 
A key difference is that SRH and SRL relax RSC and RSS. Indeed, the former pair of conditions entail only local constraints, while the latter pair entail global constraints.

The case of criterion (c), which ensures the stability of the support, is more complicated.
One can easily concoct a scenario in which $\| \bbeta_{*} \|_0 = \| \bbeta_{k} \|_0$ and 
$\| \bbeta_{k} - \bbeta_{*} \|_2$ can be made arbitrarily small, but $\supp(\bbeta_{*}) \neq \supp(\bbeta_{k})$. 
Recent research \cite{yuan2016exact} directly addresses criterion (c) by requiring that the smallest nonzero entry $\beta_{\text{min}}$ of $\bbeta_{*}$ exceed 
$\| \nabla f(\bbeta_{*}) \|_{\infty}$. While a notable achievement, this result is of mainly academic
interest since $\bbeta_{*}$ is rarely known in advance. Taken together, the results for
the three convergence criteria suggest that IHT convergence is reasonably reliable for GWAS. 
In this context, we expect that IHT will recover the true model provided that the SNP markers are not in strong linkage disequilibrium and the magnitudes of the true regression
coefficients are not too small.

\subsubsection{IHT step sizes}

Computing a reasonable step size $\mu$ is important for ensuring stable descent in projected gradient schemes.
For the case of least squares regression,
our implementation of IHT uses the ``normalized"  update of Blumensath and Davies \cite{blumensathdavies2010}.
At each iteration $m$, this amounts to employing the step size 
\begin{equation*}
	\mu_{m} = \frac{\| \bbeta^{m} \|_2^2}{\| \bX \bbeta^{m} \|_2^2}.
\end{equation*}
Convergence is guaranteed provided that $\mu_m < \omega_m$,
where
\begin{equation}
	\omega_m = (1 - c) \frac{ \left\| \bbeta^{m+1} - \bbeta^{m} \right\|_2^2 }{ \left\| \bX(\bbeta^{m+1} - \bbeta^{m} \right)\|_2^2 }
\end{equation}
for some constant $0 < c \ll 1$.
One can interpret $\omega$ as the normed ratio of the difference between successive iterates versus the difference between successive estimated responses.

\subsubsection{Bandwidth optimization of IHT}

Analysis of large GWAS datasets requires intelligent handling of memory and read/write operations. 
Our software reads datasets in PLINK binary format. 
The PLINK compression protocol stores each genotype in two bits of a 64-bit float,
thus achieving 32x compression. 
Although PLINK compression facilitates storage and transport of data, it complicates linear algebra operations. 
On small datasets, we store the design matrix $\bX$ in floating point. 
On large datasets, we store both a compressed $\bX$ and a compressed transpose $\bX^T\! .$
The transpose $\bX^T$ is used to compute the gradient \eqref{normal_grad},
while $\bX$ is used to compute the predicted response $\bX \bbeta$.
The counterintuitive tactic of storing both $\bX$ and $\bX^T$ roughly doubles the memory required to store genotypes.
However, it facilitates accessing all data in column-major and unit stride order, thereby ensuring that all linear algebra operations maintain full memory caches. 

Good statistical practice dictates standardizing all predictors; otherwise, parameters are penalized non\-uniformly.
Standardizing nongenetic covariates is trivial.
However, one cannot store standardized genotypes in PLINK binary format.
The remedy is to precompute and cache vectors $\bu$ and $\bv$ containing the mean and 
inverse standard deviation, respectively, of each of the $p$ SNPs. The product $\bX_{\text{st}}\bbeta$ invoking the standardized predictor matrix $\bX_{\text{st}}$ can be recovered via the formula
\begin{equation*}
    \bX_{\text{st}}\bbeta = \bX \diag(\bv)\bbeta - \bone \bu^T \diag(\bv) \bbeta,
\end{equation*} 
where $\bone$ is an $n$-vector of ones and $\diag(\bv)$ is a diagonal matrix with $\bv$ on the main diagonal.
Thus, there is no need to explicitly store $\bX_{\text{st}}$. 

On-the-fly standardization is a costly operation and must be employed judiciously. 
For example,
to calculate $\bX \bbeta$ we exploit the structural sparsity of $\bbeta$ by decompressing and standardizing just the submatrix $\bX_{k}$ of $\bX$ corresponding to the $k$ nonzero values in $\bbeta$.
We then use $\bX_{k}$ for parameter updates until we observe a change in the support of $\bbeta$. 
Unfortunately, calculation of the gradient $\nabla f(\bbeta)$ offers no such optimization because it requires a fully decompressed matrix $\bX^T\! .$
Since we cannot store all $n \times p$ standardized genotypes in floating point format,
the best that we can achieve is standardization on the fly every time we update the gradient.

\subsubsection{Parallelization of IHT}

Our implementation of IHT for PLINK files relies on two parallel computing schemes.
The first makes heavy use of multicore computing with shared memory arrays to distribute computations over all cores in a CPU.
For example, suppose that we wish to compute in parallel the column means of $\bX$ stored in a shared memory array.
The mean of each column is independent of the others, so the computations distribute naturally across multiple cores.
If a CPU contains four available cores, then we enlist four cores for our computations, one master and three workers.
Each worker can see the entirety of $\bX$ but only works on a subset of its columns.
The workers compute the column means for the three chunks of $\bX$ in parallel.
Column-wise operations, vector arithmetic, and matrix-vector operations fit within this paradigm.

The two most expensive operations are the matrix-vector multiplications $\bX \bbeta$ and $\bX^T (\by - \bX \bbeta)$.
We previously discussed intelligent computation of $\bX \bbeta$ via $\bX_{k} \bbeta_{k}$. 
Dense multithreaded linear algebra libraries such as BLAS facilitate efficient computation of $\bX_{k} \bbeta_{k}$.
Consequently, we obtain $\bX \bbeta$ in $\mathcal{O}(nk)$ total operations.
In contrast, the gradient criterion \eqref{normal_grad} requires a completely dense matrix-vector multiplication with a run-time complexity of $\mathcal{O}(np)$.
We could lighten the computational burden by cluster computing,
but communication between the different nodes then takes excessive time. 

A reasonable alternative for acceleration is to calculate the gradient on a GPU running the OpenCL kernel code. 
An optimal GPU implementation must minimize memory transactions between the device GPU and the host CPU.
Our solution is to push the compressed PLINK matrix $\bX$ and its column means and precisions onto the device at the start of the algorithm.
We also cache device buffers for the residuals and the gradient.
Whenever we calculate the gradient,
we compute the $n$ residuals on the host and then push the residuals onto the device.
At this stage,
the device executes two kernels.
The first kernel initializes many workgroups of threads and distributes a block of $\bX^T (\by - \bX \bbeta)$ to each workgroup. 
Each thread handles the decompression, standardization, and computation of the
inner product of one column of $\bX$ with the residuals.
The second kernel reduces across all thread blocks and returns the $p$-dimensional gradient.
Finally, the host pulls the $p$-dimensional gradient from the device.
Thus, after the initialization of the data, our GPU implementation only requires the host and device to exchange $p + n$ floating point numbers per iteration.

\subsection{Model selection \label{model_selection_section}}

Given a regularization path computed by IHT, the obvious way to choose the best model along the path
is to resort to simple $q$-fold cross-validation with mean squared error (MSE) as selection criterion. 
For a path of user-supplied model sizes $k_1, k_2, \ldots, k_r$, our implementation of IHT fits the entire path on the $q - 1$ training partitions. 
We then view the $q$th partition as a testing set and compute its mean squared error (MSE).
Finally, we determine the model size $k$ with minimum MSE and refit the data subject to $\| \bbeta \|_0 \leq k$.

\section{Results}
\label{sec:results}

We tested IHT against LASSO and MCP on data from the Northern Finland Birth Cohort 1966 (NFBC1966) \cite{sabattietal2009}.
These data contain several biometric phenotypes for 5402 patients genotyped at 370,404 SNPs.
We imputed the missing genotypes in $\bX$ with Mendel \cite{ayerslange2008} and performed quality control with PLINK 1.9 beta \cite{changetal2015}. 
Our numerical experiments include both simulated and measured phenotypes.
For our simulated phenotype, we benchmarked the model recovery and predictive performance of our software against \texttt{glmnet} v2.0.5 and \texttt{ncvreg} v3.6.0 \cite{brehenyhuang2011, friedmanetal2010} in R v3.2.4. The statistical analysis summarized in
Sections \ref{sec:speedrun} and \ref{sec:gwas} includes as nongenetic covariates the \textsc{SexOCPG} factor,
which we calculated per Sabatti et al., 
and the first two principal components of $\bX$,
which we calculated with PLINK 1.9.
All numerical experiments  were run on a single compute node equipped with four 6-core 2.67Ghz Intel Xeon CPUs and two NVIDIA Tesla C2050 GPUs, each with 6Gb of memory. 
To simulate performance on a workstation, the experiment only used one GPU and one CPU. 
The computing environment consisted of 64-bit Julia v0.5.0 with the corresponding OpenBLAS library and LLVM v3.3 compiler. 

\subsection{Simulation}
\label{sec:simulation}

The goal of our first numerical experiment was to demonstrate the superior model selection performance of IHT versus LASSO and MCP.
Here we used only the matrix $\bX_{\text{chr1}}$ of 23,695 SNPs from chromosome 1 of the NFBC1966 dataset. 
This matrix is sufficiently small to render PLINK compression and GPU acceleration unnecessary.
$\bX_{\text{chr1}}$ uses the 5289 cases with observed BMI.
Note that this number is larger than what we will use in Sections \ref{sec:speedrun} and \ref{sec:gwas}; no exclusion criteria were applied here since the phenotype was simulated. 
We standardized observed genotype dosages and then set unobserved dosages to 0.
    We simulated $\bbeta_{\text{true}}$ for true model sizes $k_{\text{true}} \in \{100, 200, 300\}$ and effect sizes independently drawn from the normal distribution $N(0,0.01)$. The simulated phenotypes were then formed by taking $\by = \bX_{\text{chr1}} \bbeta_{\text{true}} + \bepsilon$,
where each component $\epsilon_i \sim N(0,0.01)$. We will refer to this
simulation scenario as having a signal-to-noise ratio (SNR) of 100\%.
To test more noisy scenarios, we simulated $\bbeta_{\text{true}}$ under three additional SNRs of 50\%, 10\%, and 5\%.
These successively lower SNRs were obtained by drawing each causal 
$\beta_{j}$ from $N(0,0.01/s)$, where $s = 2, 10, 20$, respectively.
We approximated the (narrow-sense) heritability $h^2$ for each combination of $\by$ and $\bbeta_{\text{true}}$ by taking the ratio of the sample
variance of the predicted vector $\bX_{\text{chr1}} \bbeta_{\text{true}}$ to the sample variance of the response vector $\by$.
To assess predictive performance,
we separated 289 individuals as a testing set and used the remaining 5000 individuals
for 5-fold cross-validation.
We generated 10 different models for each $k_{\text{true}}$.
For each replicate, we ran regularization paths of 100 model sizes $k_0, k_0 + 2, \ldots, k_{\text{true}}, k_{\text{true}} + 2, \ldots, k_0 + 200$ straddling $k_{\text{true}}$ and chose the model with minimum MSE.

\begin{figure}
    \centering
    \begin{subfigure}[t]{0.49\textwidth}
        \includegraphics[width=\linewidth]{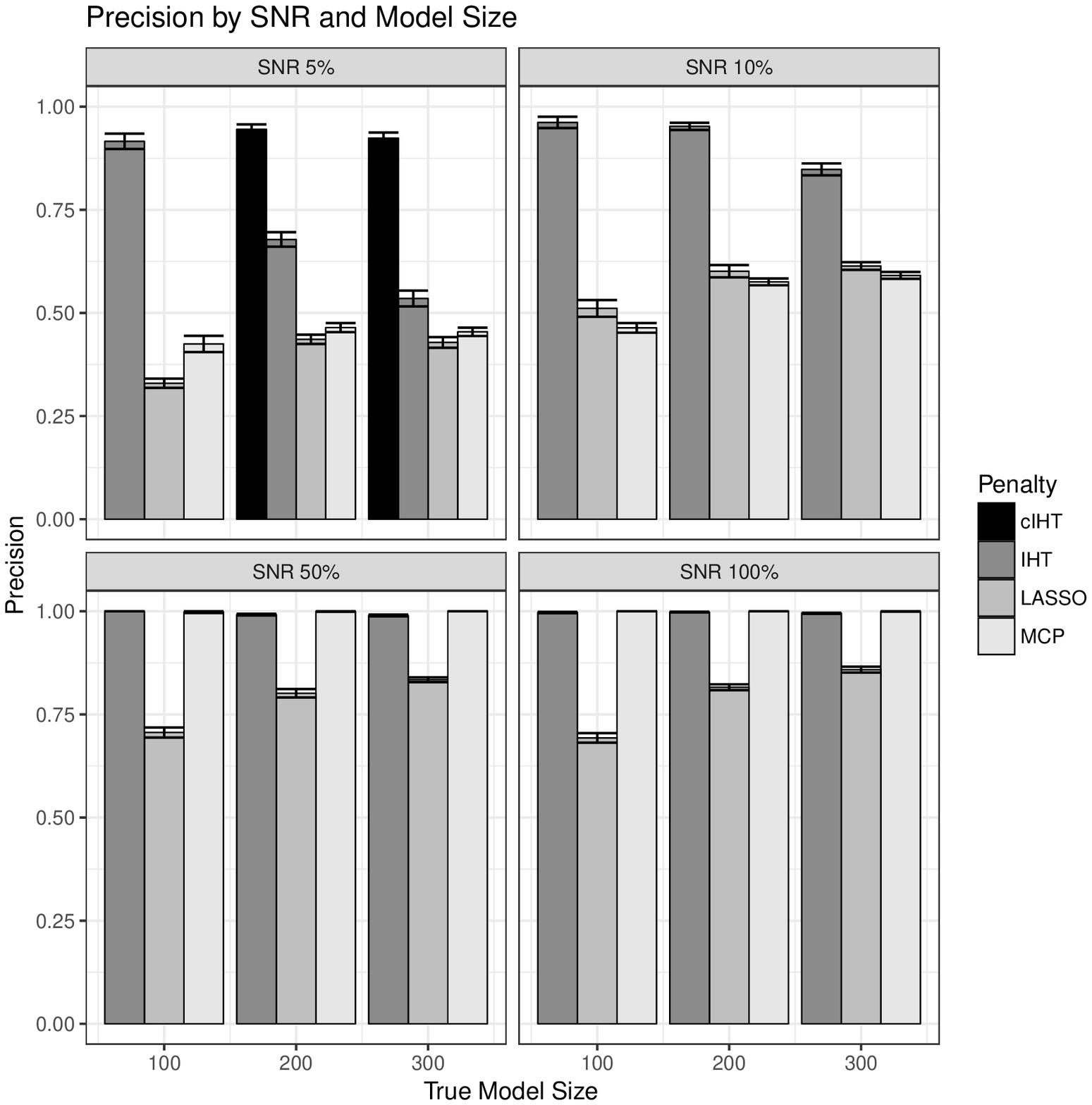}
        %\caption{Precision estimates from simulations, stratified by SNRs, model size, and algorithms.}
        \label{fig:sim_precison}
    \end{subfigure}
    \hfill
    \begin{subfigure}[t]{0.49\textwidth}
        \includegraphics[width=\linewidth]{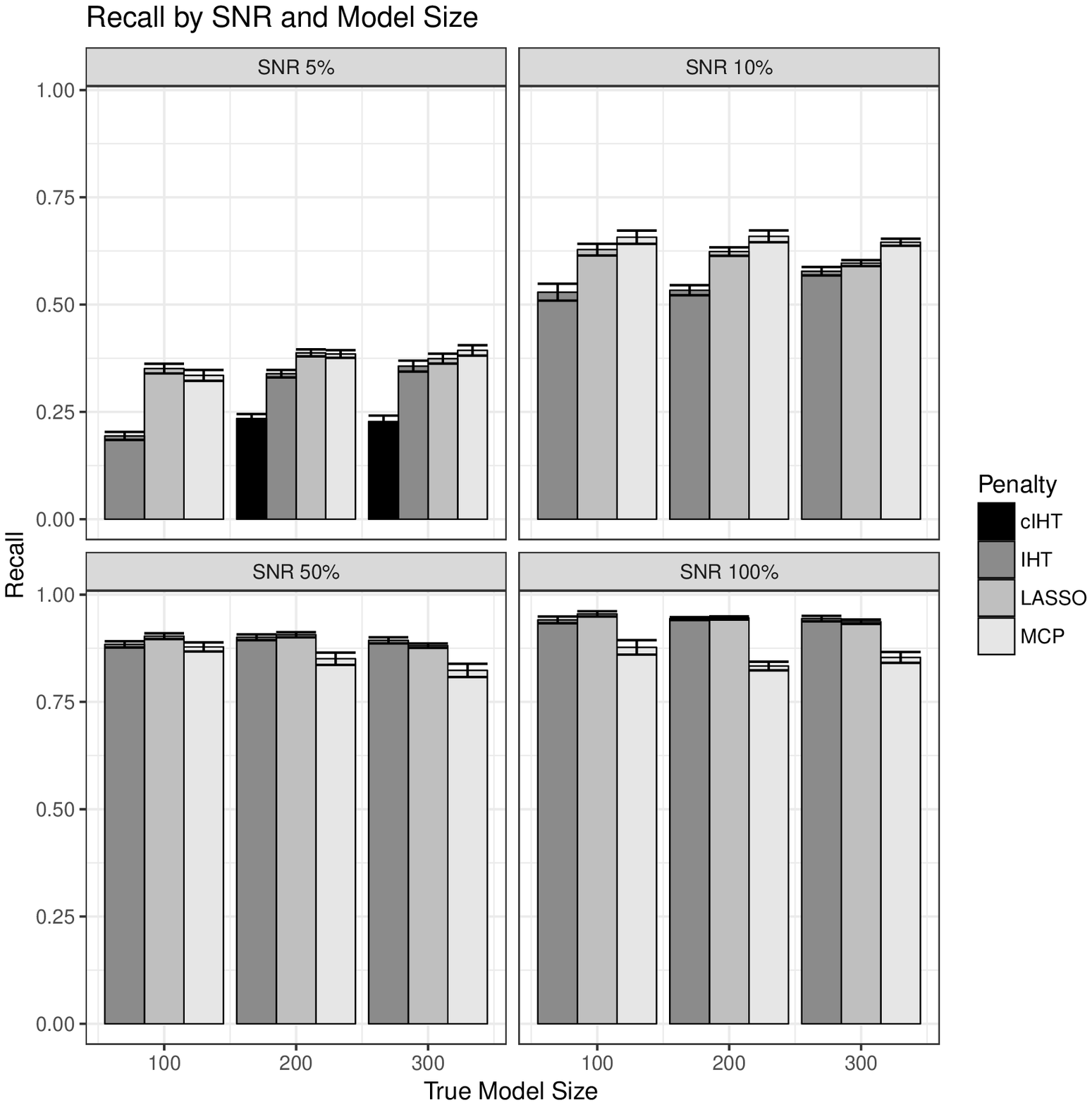}
        %\caption{Recall estimates from simulations, stratified by SNRs, model size, and algorithms.}
        \label{fig:sim_recall}
    \end{subfigure}
    \vspace{0.25in}
    \begin{subfigure}[t]{0.49\textwidth}
        \includegraphics[width=\linewidth]{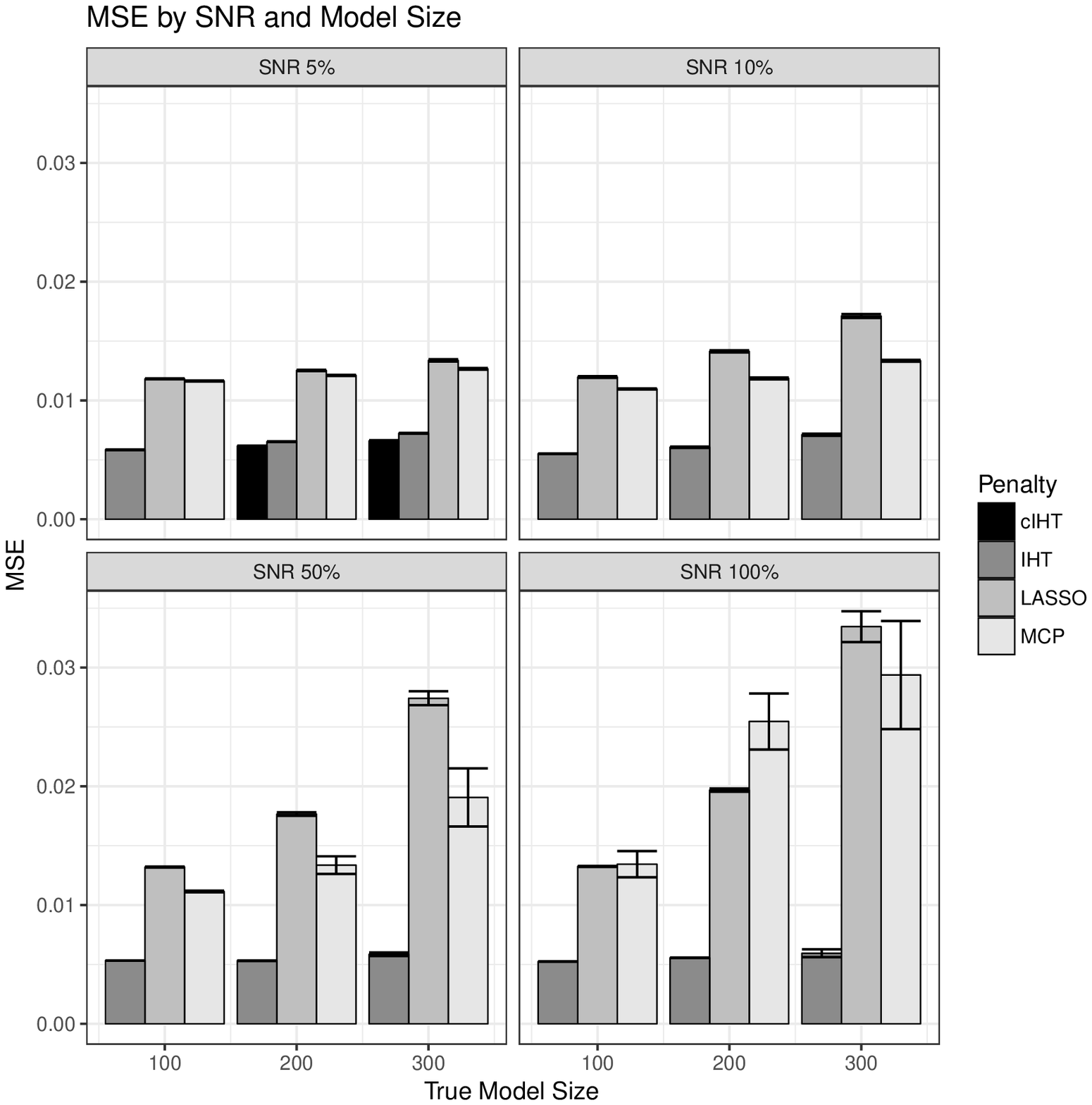}
        %\caption{Mean-squared error estimates from simulations, stratified by SNRs, model size, and algorithms.}
        \label{fig:sim_mse}
    \end{subfigure}
    \hfill
    \begin{subfigure}[t]{0.49\textwidth}
        \includegraphics[width=\linewidth]{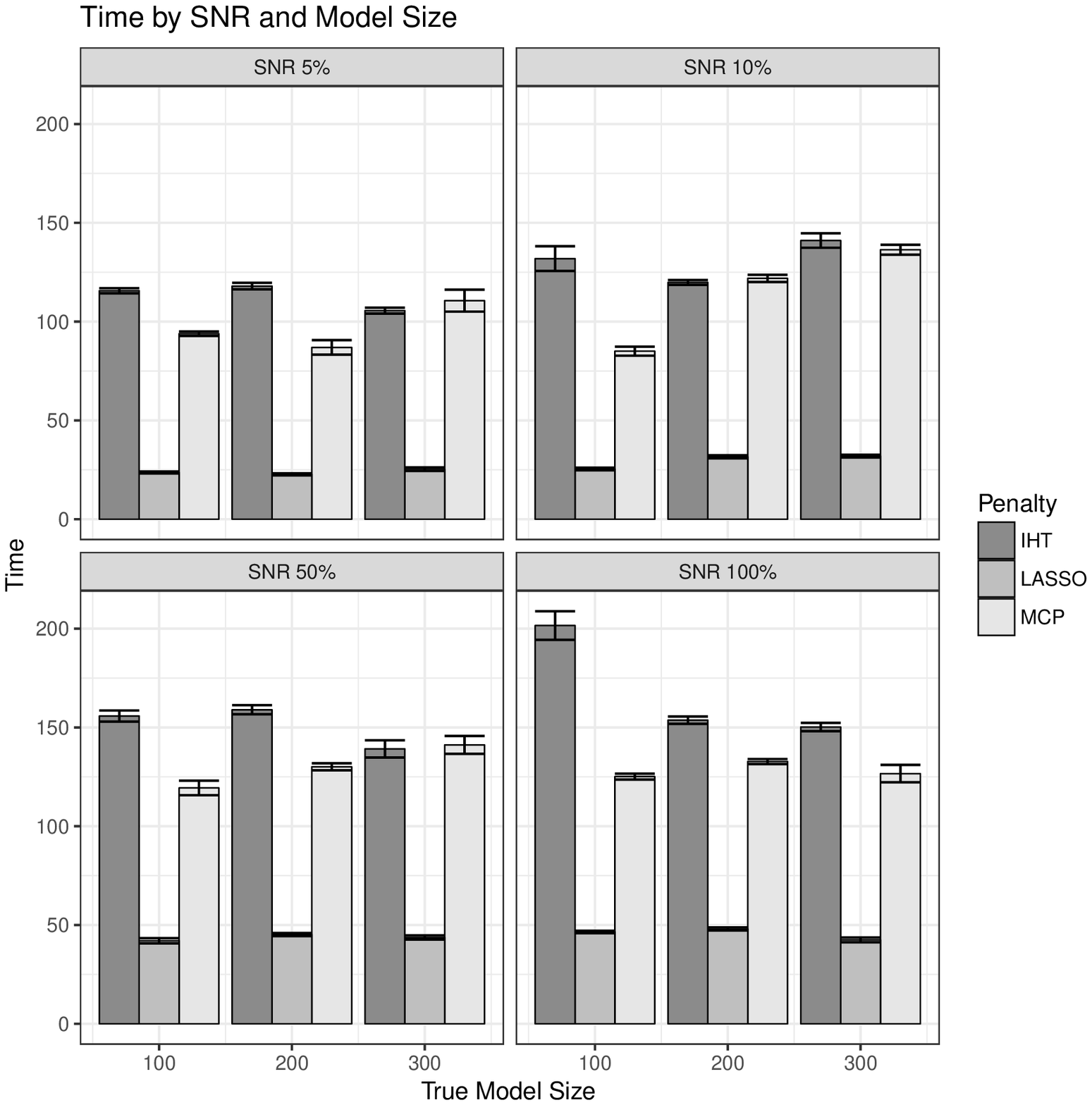}
        %\caption{Compute time estimates from simulations, stratified by SNRs, model size, and algorithms.}
        \label{fig:sim_time}
    \end{subfigure}
    %\caption{Results from the simulations for each of the three algorithms, stratified by SNR and true model size and averaged over 10 simulation replicates. (a) Precision, (b) Recall, (c)  Mean-squared error, and (d) Compute time for each simulation scenario.}
    \caption{}
    \label{fig:sims}
\end{figure}

The cross-validation choice of model size is straightforward under IHT.
For cross-validation with LASSO and MCP,
we used the cross-validation and response prediction routines in \texttt{glmnet} and \texttt{ncvreg}. 
To ensure roughly comparable lengths of regularization paths and therefore commensurate compute times,
we capped the maximum permissible degrees of freedom at $k_{\text{true}} + 100$ for both LASSO and MCP regression routines. 
The case of MCP regression is peculiar since \texttt{ncvreg} does not cross-validate the $\gamma$ parameter.
We modified the approach of \cite{brehenyhuang2011} to obtain a suitable $\gamma$ for each model.
Their protocol entails cross-validating $\lambda$ once with the default $\gamma = 3$ and checking if the optimal $\lambda$, which we call $\lambda_{\text{best}}$, exceeds the minimum lambda $\lambda_{\text{min}}$ guaranteeing a convex objective. 
Whenever $\lambda_{\text{best}} \leq \lambda_{\text{min}}$, we incremented $\gamma$ by $1$ and cross-validated $\lambda$ again.
We repeated this process until $\lambda_{\text{best}} > \lambda_{\text{min}}$.
The larger final $\gamma$ then became the default for the next simulation,
thereby amortizing the selection of a proper value of $\gamma$ across all 10 simulations for a given $k_{\text{true}}$. 
This procedure for selecting $\gamma$ ensured model selection stability while simultaneously avoiding expensive cross-validation over a full grid of $\gamma$ and $\lambda$ values. 
The reported compute times for MCP reflect this procedure, though we never needed to increment $\gamma$ in our simulations.

Figure \ref{fig:sims} shows simulation results for precision, recall, prediction error (MSE), and compute time for each SNR and each model size $k_{\text{true}}$.
Here we compute precision as the ratio $k_{\text{recovered}} / k$ of recovered causal markers $k_{\text{recovered}}$ to the number of recovered markers $k$. 
Recall is computed as the ratio $k_{\text{recovered}} / k_{\text{true}}$ of recovered causal markers to the true number $k_{\text{true}}$ of causal markers.
We see that IHT consistently maintains good precision and superior prediction error (MSE), even with decreasing SNR.
At high SNR, its precision is better than that of LASSO, while its recall is better than MCP.
As SNR decreases, IHT cedes its advantage in recall.
Despite these benefits, IHT pays only a modest price in computational speed versus LASSO and MCP.
For example, IHT is only 4-6 times slower than LASSO, and it is often competitive with MCP in speed.
We note that \texttt{glmnet} often exited before testing all 100 values of $\lambda$ on its regularization path,
so the timing values do not constitute truly rigorous performance comparisons.

Careful readers may observe that the precision of IHT suddenly declines for $k = 200$ and $k = 300$ for SNR 5\%.
In these scenarios, IHT returns the minimum model size of the regularization path; for $k = 200$, the minimum is 100, while for $k = 300$ the minimum is 200.
We suspected that this behavior was an artifact of our simulation scenario.
Indeed, by testing every model $k = 1, 2, \ldots, k_{\text{true}} + 100$, for these scenarios, we found that IHT estimated models sizes smaller than our simulation setup originally allowed. 
These results appear in Figure \ref{fig:sims} for SNR 5\% and $k = 200, 300$ as ``cIHT'' for ``corrected IHT''.
We discarded the timing results in this case since they are not comparable to our previous results.
The corrected IHT results show IHT regaining its edge in precision over the other algorithms.

\begin{figure}
    \includegraphics[width=\linewidth]{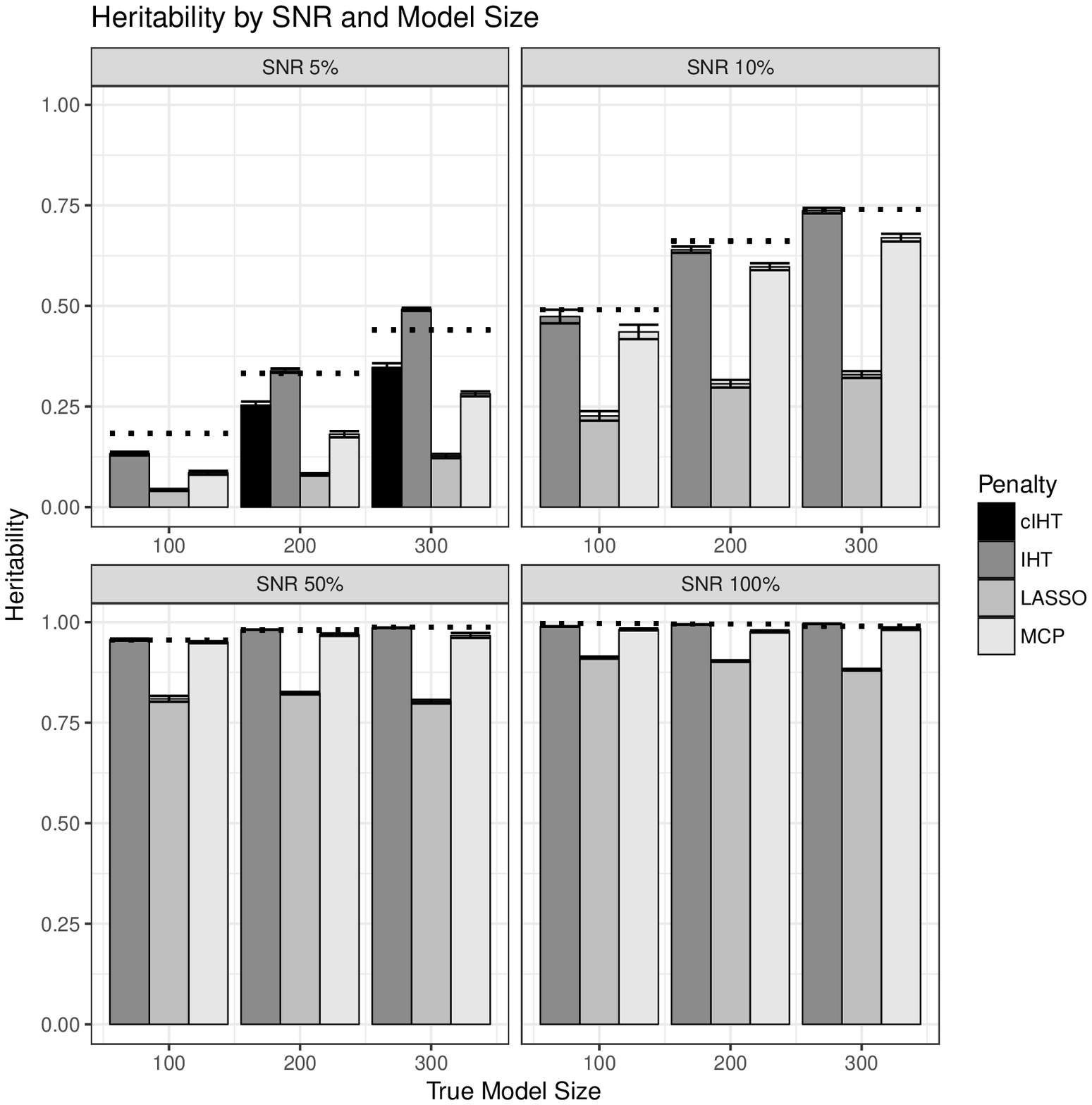}
    %\caption{Estimates of narrow-sense heritability for IHT, LASSO, and MCP, stratified by SNR and true model size, averaged across 10 simulation replicates.}}
    \caption{}
    \label{fig:h2}
\end{figure}

As noted previously, the heritability of the true model is given by $h^2_{\text{true}} = \Var(\bX_{\text{chr1}} \bbeta_{\text{true}} ) / \Var(\by)$.
For each model $\bbeta^*$ selected by IHT, LASSO, or MCP, we obtain the estimated heritability $h_{\text{estimated}}^2$ similarly by substituting
 $\bbeta^*$ for $\bbeta_{\text{true}}$. 
Figure \ref{fig:h2} shows the heritability estimates from our simulation.
In each case, we average the estimated heritability over all 10 simulation runs for each algorithm, model size, and SNR.
A dotted line demarcates $h_{\text{true}}^2$ for each scenario, also averaged over all 10 simulation runs.
We can see that as SNR decreases, IHT consistently estimates heritability better than either LASSO or MCP.
For SNR 5\% and $k = 200, 300$, the IHT estimates of $h^2$ exceed $h_{\text{true}}^2$ and are obviously unrealistic.
Over-estimation stems from our failure to allow for sufficiently small model sizes.
Correcting this mistake eliminates the anomalies in estimated $h^2$ and shows that IHT still estimates $h^2$ better than the two competing algorithms in noisy scenarios.

\subsection{Analysis of compressed data with IHT}
\label{sec:speedrun}

Our next numerical experiment highlights the sacrifice in computational speed that IHT incurs with compressed
genotypes. The genotype matrix $\bX_{\text{chr1}}$ is now limited to
patients with both BMI and weight directly observed, a condition imposed by  \cite{sabattietal2009}. 
The response vector $\by$ is the log body mass index (BMI) from NFBC1966.
As mentioned previously, we included the \textsc{SexOCPG} factor and the first two principal components as nongenetic covariates. 
We then ran three different schemes on a single compute node. 
The first used the floating point version of $\bX_{\text{chr1}}$.
We did not explicitly enable any multicore calculations.
For the second run,
we used a compressed copy of $\bX_{\text{chr1}}$ with multicore options enabled,
but we disabled the GPU acceleration.
The third run used the compressed $\bX_{\text{chr1}}$ data with both multicore and GPU acceleration.
We ran each algorithm over a regularization path of model sizes $k = 5, 10, 15, \ldots, 100$ and averaged compute times over 10 runs. 
For all uncompressed arrays, we used double precision arithmetic.

Table \ref{tab:speedrun} shows the compute time statistics.
The floating point variant is clearly the fastest, requiring about 8 seconds to compute all models. The analysis using PLINK compression with a multicore CPU suffers a 17x slowdown,
clearly demonstrating the deleterious effects of repeated decompression and on-the-fly standardization. 
Enabling GPU acceleration leads to an 8x slowdown and fails to entirely bridge the performance gap imposed by compressed data.
In defense of GPU computing, it is helpful to make a few remarks.
First, the computational burden of analyzing just a single chromosome does not fully exploit GPU capabilities; hence, Table \ref{tab:speedrun} does not demonstrate the full potential of GPU acceleration.
Second, the value of GPU acceleration becomes more evident in cross-validation: a 5-fold cross-validation on one machine requires either five hexcore CPUs or one hexcore CPU and one GPU.
The latter configuration lies within modern workstation capabilities. 
Third, our insistence on the use of double precision arithmetic dims the luster of GPU acceleration. Indeed, in our experience using compressed arrays and a GPU with single precision arithmetic is only 1.7x slower than the corresponding floating point compute times. 
Furthermore, while we limit our computations to one CPU with six physical cores, including additional physical cores improves compute times even for compressed data without a GPU. 

\begin{table}[!t]
    \begin{center}
        \begin{tabular}{lcc}
            \toprule
            Data type                       & Mean Time & Standard Deviation \\
            \midrule
            Uncompressed Data               &   8.27   &   0.056 \\
            Compressed Data, no GPU         &   141.28 &   0.998 \\
            Compressed Data with CPU + GPU  &   65.48  &   0.040 \\
            \bottomrule
        \end{tabular}
        \caption{Computational times in seconds on NFBC1966 chromosome 1 data.} 
        \label{tab:speedrun}
    \end{center}
\end{table}

\subsection{Application of IHT to lipid phenotypes}
\label{sec:gwas}

For our final numerical experiment, we embarked on a genome-wide search for associations based on the full $n \times p$ NFBC1966 genotype matrix $\bX$. 
In addition to BMI, this analysis considered three additional phenotypes from \cite{sabattietal2009}: HDL cholesterol (HDL), LDL cholesterol (LDL), and triglycerides (TG).
HDL, LDL, and TG all rely on \textsc{SexOCPG} and the first two principal components as nongenetic covariates.
Quality control on SNPs included filters for minor allele frequencies below $0.01$ and Hardy-Weinberg 
$P$-values below $10^{-5}$. Subjects with missing traits were excluded from analysis. 
We applied further exclusion criteria per Sabatti et al. \cite{sabattietal2009};
for analysis with BMI, we excluded subjects without direct weight measurements,
and for analysis of TG, HDL, and BMI, we excluded subjects with nonfasting blood samples and subjects on diabetes medication.
These filters yield different values of $n$ and $p$ for each trait.  Table \ref{tab:counts} records problem dimensions and trait transforms.
To select the best model size, we performed 5-fold cross-validation over a path of sparsity levels $k = 1, 2, \ldots, 50$.
Refitting the best model size yielded effect sizes.
Table \ref{tab:counts} records the compute times and best model sizes, while Table \ref{tab:gwasresults} shows the SNPs recovered by IHT.

\begin{table}[!t]
    \begin{center}
    \begin{tabular}{cccccc}
        \toprule
        Phenotype & $n$  & $p$ & Transform & $k_{\text{best}}$ & Compute Time (Hours) \\
        \midrule
        BMI       & 5122 & 333,656 & log  & 2  & 1.12 \\
        HDL       & 4729 & 333,774 & none & 9  & 1.28 \\
        LDL       & 4715 & 333,771 & none & 6  & 1.32 \\
        TG        & 4728 & 333,769 & log  & 10  & 1.49 \\
        \bottomrule
    \end{tabular}
    \caption{Dimensions of data used for each phenotype in GWAS experiment.
             Here $n$ is the number of cases, $p$ is the number of predictors (genetic + covariates), and $k_{\text{best}}$ is the best cross-validated model size.
             Note that $k_{\text{best}}$ includes nongenetic covariates.} 
    \label{tab:counts}
    \end{center}
\end{table}

One can immediately see that IHT does not collapse causative SNPs in strong linkage disequilibrium.
IHT finds the adjacent pair of SNPs rs6917603 and rs9261256 for HDL.
For TG, rs11974409 is one SNP separated from rs2286276, while SNP rs676210 is one SNP separated from rs673548.
Note that rs673548 is not in Table \ref{tab:counts} since IHT does not flag rs673548.
However, its association with TG in NFBC1966 is known. \cite{sabattietal2009}
Common sense suggests treating each associated pair of SNPs as a single predictor.

\begin{table}[!t]
    \begin{center}
    \begin{tabular}{cccccl}
        \toprule
        Phenotype & Chromosome & SNP & Position & $\beta$ & Status \\ 
        \midrule
        BMI & 6  & rs6917603  & 30125050  & -0.01995 & Unreported \\
        \hline
        HDL & 6  & rs6917603  & 30125050  &  0.10100 & Reported \\
            & 6  & rs9261256  & 30129920  & -0.06252 & Nearby   \\
            & 11 & rs7120118  & 47242866  & -0.03351 & Reported \\
            & 15 & rs1532085  & 56470658  & -0.04963 & Reported \\
            & 16 & rs3764261  & 55550825  & -0.02808 & Reported \\
            & 16 & rs7499892  & 55564091  &  0.02625 & Reported \\
        \hline
        LDL & 1  & rs646776   & 109620053 &  0.09211 & Reported \\
            & 2  & rs693      & 21085700  & -0.08544 & Reported \\
            & 6  & rs6917603  & 30125050  & -0.07536 & Reported \\
        \hline
         TG & 2  & rs676210   & 21085029  &  0.03633 & Nearby     \\
            & 2  & rs1260326  & 27584444  & -0.04088 & Reported   \\
            & 6  & rs7743187  & 25136642  &  0.03450 & Unreported \\
            & 6  & rs6917603  & 30125050  & -0.08215 & Unreported \\
            & 7  & rs2286276  & 72625290  &  0.01858 & Reported   \\
            & 7  & rs11974409 & 72627326  &  0.01759 & Nearby     \\
            & 8  & rs10096633 & 19875201  &  0.03781 & Reported   \\
            & 13 & rs3010965  & 60937883  &  0.02828 & Unreported \\
            & 19 & rs2304130  & 19650528  &  0.03039 & Reported   \\
        \bottomrule
    \end{tabular}
    \caption{Computational results from the GWAS experiment. Here $\beta$ is the calculated effect size. Known associations include the relevant citation.} 
    \label{tab:gwasresults}
    \end{center}
\end{table}

Our analysis not only replicates several associations from the literature but also
finds new ones as well. For example, \citeauthor{sabattietal2009} found associations between TG and the SNPs rs1260326 and rs10096633, while rs2286276 was identified elsewhere.
The SNPs rs676210, rs7743187, rs6917603, and rs3010965 are new associations with TG.
We find that SNP rs6917603 is associated with all four traits; the BMI connection was missed by \citeauthor{sabattietal2009}. 

The association of SNP rs6917603 with BMI may be secondary to its association to lipid levels. 
Indeed, a multivariate analysis of the combined traits
BMI, HDL, LDL, and TG reveals a much stronger association to rs6917603 than BMI alone does. In the former case $P = 9.88 \times 10^{-105}$ while in the latter case
$P = 1.19 \times 10^{-15}$.
The four traits exhibit fairly high correlations, with the correlation coefficient 
$\rho$ ranging between $0.25$ and $0.35$.  We conjecture that some of the observed pleiotropic effect of rs6917603 may be explained by these trait correlations.
A more extensive analysis that incorporates nearby genetic markers in LD may clarify the association pattern displayed by these correlated traits \cite{fan2013functional, fan2015gene, wang2015pleiotropy}. 
Note that the aforementioned $P$-values are meant to be compared to each other, not to previous GWAS associations.
Our pleiotropic analysis of rs6917603 makes no assertion of a genome-wide significant association for any of the phenotypes since the $P$-values reported here are likely inflated by selection bias \cite{taylortibshirani2015}.

IHT flags an association between SNP rs2304130 and TG. 
This association was validated in a large meta-analysis of 3,540 cases and 15,657 controls performed after \cite{sabattietal2009} was published. 
Finally, some of the effect sizes in Table \ref{tab:gwasresults} are difficult to interpret.
For example, IHT estimates effects for rs10096633 ($\beta = 0.03781$) and rs1260326 ($\beta = -0.04088$) that are both smaller and in opposite sign to the estimates in \cite{sabattietal2009}. 

%% figure included, but caption sent to end? 
\begin{figure}
    \includegraphics[width=0.9\textwidth]{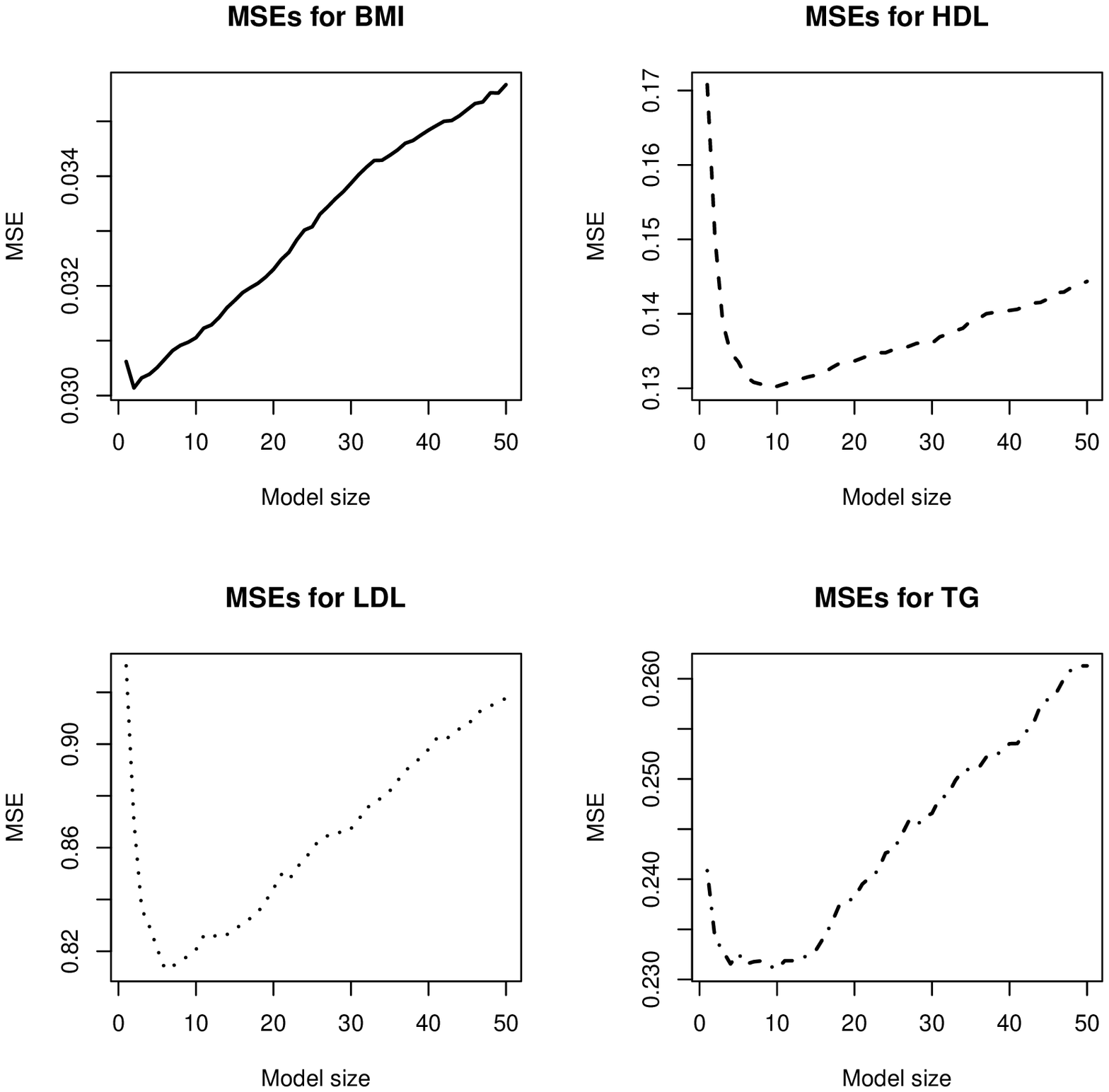}
    %\caption{Mean squared error as a function of model size, as averaged over 5 cross-validation slices, for four lipid phenotypes from NFBC 1966.}
    \caption{}
    \label{fig:stampeed_mses}
\end{figure}

The potential new associations of TG with SNPs rs7743187 and rs3010965 are absent from the literature. 
Furthermore, our analysis misses borderline significant associations identified by Sabatti et al \cite{sabattietal2009},
such as rs2624265 for TG and rs9891572 for HDL; these SNP associations went unreplicated in later studies.
In this regard it is worth emphasizing that the best model size $k_{\text{best}}$ delivered by cross-validation is a guide rather than definitive truth. 
Figure \ref{fig:stampeed_mses} shows that the difference in MSE between $k_{\text{best}}$ and adjacent model sizes can be quite small. 
Models of a few SNPs more or less than $k_{\text{best}}$ predict trait values about as well. 
Thus, TG with $k_{\text{true}} = 10$ has MSE = 0.2310, while TG with $k_{\text{true}} = 4$ has MSE = 0.2315. 
Refitting the TG phenotype with $k = 4$ yields only three significant SNPs: rs1260326, rs6917603, and rs10096633.
The SNPs rs7743187 and rs3010965 are absent from this smaller model, so we should be cautious in declaring new associations.
This example also highlights the value in computing many model sizes,
which univariate regression schemes typically overlook.

In light of the simulation results from Section \ref{sec:simulation}, an obvious question is how our IHT results on real data compare to those from LASSO and MCP. 
To this end, we ran LASSO and MCP on each of the four phenotypes and the full genotype matrix.
For LASSO, we used PLINK 1.9 \cite{changetal2015} after tuning the regularization path using heritability estimates from GCTA \cite{yangetal2010,yang2011gcta}.
For MCP, we used the GWAS module in Mendel \cite{langeetal2013,wu2009genome,wulange2008,zhou2010association}.
Since Mendel requires the user to specify the desired model size $k$, we simply used the best cross-validated $k$ from IHT for each phenotype. 
A rigorous comparison of model selection performance would use a predictive measure such as out-of-sample predictive accuracy.
Unfortunately, neither PLINK or Mendel return MSEs, so we cannot compare predictive power directly.
Instead, we compare to which extent the marker sets selected by IHT agree with corresponding sets selected by LASSO or MCP, similar in spirit to our previous validation of IHT results with the literature.
Tables showing the markers recovered by LASSO and MCP appear as supplementary material.
In general, LASSO returns a superset of the markers selected by IHT, as expected from our simulations.
MCP and IHT usually but not always select similar markers.
Given our limited ability to compare the three methods, the set of markers selected by IHT seems reasonable in light of results from LASSO and MCP,
but we refrain from drawing conclusions about comparative predictive performance among the three methods. 

Finally, we comment on compute times. 
IHT requires about 1.5 hours to cross-validate the best model size over 50 possible models using double precision arithmetic.
Obviously, computing fewer models can decrease this compute time substantially.
If the phenotype in question is scaled correctly,
then analysis with IHT may be feasible with single precision arithmetic,
which yields an additional speedup as suggested in Section \ref{sec:speedrun}. 
Analyses requiring better accuracy will benefit from the addition of double precision registers in newer GPU models.  
Thus, there is further room for speedups without sacrificing model selection performance.

\section{Discussion}
\label{sec:discussion}

The mathematical literature on big data analysis is arcane to the point of being nearly inaccessible to geneticists.
In addition to absorbing the obvious mathematical subtleties, readers must be wary of the hype that infects many papers.
This manuscript represents our best effort to sort through the big data literature and identify advances most pertinent to the analysis of GWAS data.
Once we decided that the iterative hard thresholding (IHT) algorithm had the greatest potential,
we set about adapting it to the needs of geneticists and comparing it to existing methods.
This paper is a synopsis of our experience in carrying out these tasks. 

Our experiments clearly demonstrate the utility of IHT in large-scale GWAS analysis.
It exhibits better model selection than more popular and mature tools such as LASSO- and MCP-regression.
Despite its nonconvex nature, IHT enjoys provable convergence guarantees under ideal regularity conditions.
We prefer IHT to other greedy algorithms because it provides the best balance of computational speed, model recovery, and convergence behavior \cite{blanchard2011phase, blanchard2015performance}.
Our software directly and intelligently handles the PLINK compression protocol widely used to store GWAS genotypes. 
Finally, IHT can be accelerated by exploiting shared-memory and massively parallel processing hardware. 
As a rule of thumb, computation times with IHT scale as $\mathcal{O}(np)$ or somewhat worse if more predictors with small effect sizes come into play. 

Lack of general support for PLINK binary genotype data, poor memory management, and primitive parallel capabilities limit the use of software such as \texttt{glmnet} and \texttt{ncvreg} in GWAS. Our IHT package \texttt{IHT.jl} enables analysis of models of any sparsity.
In contrast,  \texttt{gpu-lasso} is designed solely for very sparse models. 
Both \texttt{IHT.jl} and \texttt{gpu-lasso} cross-validate for the best model size over a range of possible models.
\texttt{IHT.jl} has the edge in scalability and model selection.
Despite these advantages, IHT is hardly a panacea for GWAS. 
Geneticists must still deal with perennial statistical issues such as correlated predictors, sufficient sample sizes, and population stratification. 
IHT can handle population stratification by inclusion of principal components as nongenetic covariates, but the onus falls upon the analyst to decide the appropriate number of PCs to use. 
Furthermore, while the estimation properties of greedy pursuit algorithms are well understood,
the theory of inference with IHT is immature \cite{yang2016selective}.
More progress has been made in understanding postselection inference with LASSO penalties \cite{lockhartetal2014,taylortibshirani2015, lee2016exact, loftustaylor2015, loftus2015selective}. 
The rapid pace of research in selective inference makes us hopeful that inference with IHT will soon become routine. 

Our analysis framework neglects the wealth of data available from whole genome sequencing. 
As sketched in Section \ref{sec:penalized_regression}, the mutual coherence condition for convergence of IHT discourages the use of IHT on strongly correlated predictors such as those that arise from analyzing genome sequences base-by-base.
Users interested in rare variant analysis could feasibly put IHT to their advantage by lumping correlated rare SNPs into univariate predictors and performing model selection on these measures of genetic burden \cite{li2008methods}.

As formulated here, the scope of application for IHT is limited to linear least squares regression. 
Researchers have begun to extend IHT to generalized linear models, particularly logistic regression \cite{bahmanirajboufounos2013, yuanlizhang2013}.
We anticipate that IHT will eventually join LASSO and MCP in the standard toolbox for sparse linear regression and sparse generalized linear regression.
In our opinion, gene mapping efforts clearly stand to benefit from application of the IHT algorithm. 

\section*{Conflict of Interest Disclosure}

The authors declare no competing interests.

\section*{Funding}

This work was supported by grants from the National Human Genome Research Institute 
[HG006139] and the National Institute of General Medical Sciences [GM053275] to K.L. and fellowship support from the National Human Genome Research Institute 
[HG002536] and the National Science Foundation [DGE-0707424] to K.L.K.

\section*{Acknowledgments}

The authors are grateful to Aditya Ramdas for his guidance on IHT and to Dennis Sun for discussions about general model selection.
We benefited from discussions about IHT at the American Institute of Mathematics.
The authors thank the two anonymous reviewers whose comments substantially improved the quality of the manuscript.
Finally, we thank the Stanford University Statistics Department for hosting us as sabbatical guests during the 2014-2015 academic year.

\bibliographystyle{natbib}
\bibliography{references.bib}
%\bibliography{iht_paper_genepi.bbl}

\end{document}